\documentclass[10pt,twocolumn,letterpaper]{article}
\usepackage{tabularx}
\usepackage{array}  
\usepackage{cvpr}              
\usepackage{algorithm}
\usepackage{algpseudocode}

\definecolor{cvprblue}{rgb}{0.21,0.49,0.74}
\usepackage[pagebackref,breaklinks,colorlinks,allcolors=cvprblue]{hyperref}

\def\paperID{}
\def\confName{CVPR}
\def\confYear{2026}

\title{Easy3E: Feed-Forward 3D Asset Editing via Rectified Voxel Flow}

\author{
    Shimin Hu \quad Yuanyi Wei \quad Fei Zha \quad Yudong Guo\thanks{Corresponding author.} \quad Juyong Zhang \\
    University of Science and Technology of China \\
    {\tt\small \url{https://ustc3dv.github.io/Easy3E/}\vspace{-1em}}
}

\begin{document}
\twocolumn[{%
\renewcommand\twocolumn[1][]{#1}%
\maketitle 
\begin{center}
    \captionsetup{type=figure}
    \includegraphics[width=\textwidth]{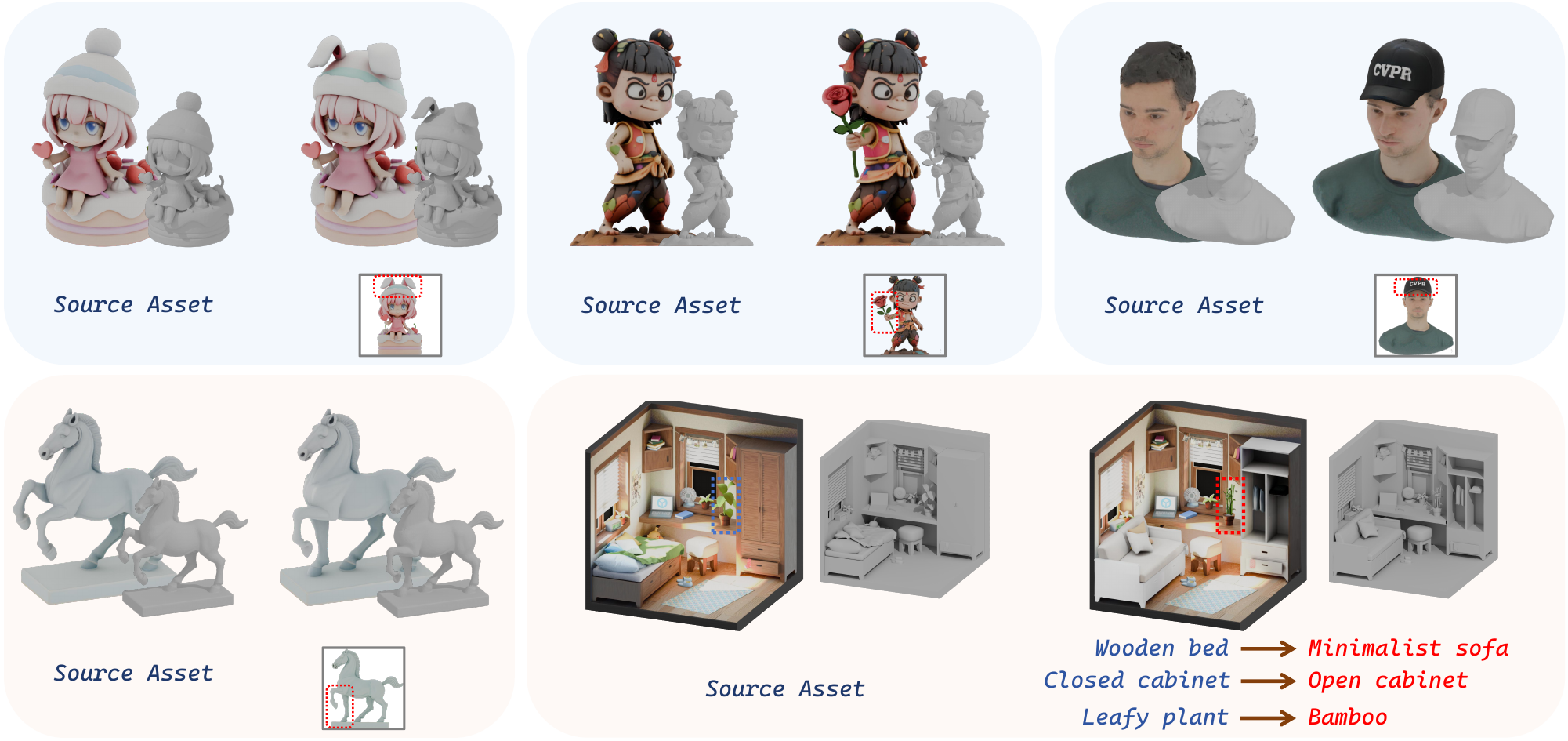} 
    \caption{We introduce Easy3E, a novel method for 3D asset editing. Guided by a single edited view and a coarse 3D mask, our method can perform both significant geometric changes and fine-grained appearance edits. Easy3E efficiently produces globally consistent and high-fidelity 3D results, demonstrating its power and flexibility across diverse assets.
}
    \label{fig:pipeline}
\end{center}
\vspace*{0.2em} 

}] 
\insert\footins{\noindent\footnotesize$^*$Corresponding Author.}
\begin{abstract}
Existing 3D editing methods rely on computationally intensive scene-by-scene iterative optimization and suffer from multi-view inconsistency. We propose an effective and feed-forward 3D editing framework based on the TRELLIS generative backbone, capable of modifying 3D models from a single editing view. Our framework addresses two key issues: adapting training-free 2D editing to structured 3D representations, and overcoming the bottleneck of appearance fidelity in compressed 3D features. To ensure geometric consistency, we introduce Voxel FlowEdit, an edit-driven flow in the sparse voxel latent space that achieves globally consistent 3D deformation in a single pass. To restore high-fidelity details, we develop a normal-guided single to multi-view generation module as an external appearance prior, successfully recovering high-frequency textures. Experiments demonstrate that our method enables fast, globally consistent, and high-fidelity 3D model editing.
\end{abstract}    
\section{Introduction}
\label{sec:intro}

3D asset editing is a fundamental task in numerous applications, such as gaming, film production, architectural visualization, and emerging fields like AR/VR and digital twins. Therefore, developing intuitive and efficient editing tools has been a long-standing challenge in computer graphics and 3D computer vision. The primary goal is to enable users to perform complex 3D modifications through simple and easy-to-use inputs, such as 2D conditions or text prompts. A framework that can translate these user inputs into coherent and high-fidelity 3D results will significantly simplify the content creation process and make 3D editing easily accessible to more users.

Existing 3D editing methods span multiple paradigms. 
Classical approaches follow the 2D-lifting pipeline~\cite{haque2023instruct, wang2024gaussianeditor,zhuang2023dreameditor}, 
where edited 2D images supervise the optimization of a 3D representation (e.g., NeRF~\cite{mildenhall2021nerf} or 3DGS~\cite{kerbl20233d}). 
Although effective for appearance-level edits, these optimization-based methods require per-scene iterative refinement and depend heavily on multi-view coverage, making them fragile when the edit introduces noticeable geometric deviation from the original asset. More recent works adopt multi-view or view-consistent diffusion models~\cite{barda2025instant3dit}, 
which improve cross-view consistency but still operate in a 2D-native feature space, requiring the model to infer 3D structure implicitly during generation. 
Such implicit reasoning limits their ability to handle edits that alter shape, topology, or volumetric occupancy, as 2D features alone provide insufficient cues for reliable 3D structural inference. Both categories rely on image-space features and therefore struggle with large geometric changes and precise structural control, especially when edits demand explicit, globally consistent manipulations of the underlying 3D structure.

In contrast, 3D-native generative models~\cite{honglrm,tang2024lgm,xiang2025structured} 
learn explicit, structured 3D latent fields directly from data. 
These representations encode geometry natively rather than reconstructing it from multiple images, 
opening a fundamentally different editing perspective: instead of optimizing a 3D scene or manipulating multi-view features, 
one can directly modify the underlying structured 3D latent space where geometry is explicitly parameterized. 
This paradigm promises feed-forward, globally coherent 3D editing, but introduces two key challenges.

First, the lack of paired 3D editing data necessitates adapting training-free 2D editing techniques~\cite{hertz2023prompt,mengsdedit,mokady2023null} to 3D latent fields. 
However, many 2D methods depend on architectural components that do not transfer to 3D, such as manipulations of cross-attention maps~\cite{hertz2023prompt,cao2023masactrl} or 2D-specific feature maps~\cite{tumanyan2023plug}. 
Second, structured 3D generative models~\cite{honglrm,xiang2025structured} utilize compact latent tokens to ensure geometric consistency and fast inference. 
This compression limits their ability to represent high-frequency texture details, 
leading to oversmoothed or low-fidelity appearance. 
Thus, the core technical questions are:  
(1) how to redesign training-free 2D editing approaches to operate on structured 3D latent spaces, and  
(2) how to restore high-quality texture details on edited geometry given limited 3D appearance priors.

To address these challenges, we propose a fully feed-forward 3D editing framework built on the TRELLIS generative backbone~\cite{xiang2025structured}. 
Given a single edited view and a user-defined editable region, our method performs both geometric and appearance editing directly in TRELLIS's sparse voxel latent space. 
We introduce Voxel FlowEdit, a latent-space editing mechanism that translates source voxels to target voxels via an adapted velocity field, enabling globally coherent geometric deformation in a single pass. 
Following this coarse transformation, we apply a structured latent repainting stage to locally refine geometry and appearance while anchoring unedited regions, ensuring consistency and detail preservation.

To overcome the limited appearance priors of 3D generative models, 
we further incorporate an optional normal-guided multi-view generative module. 
It synthesizes high-fidelity auxiliary views aligned with the edited geometry, 
providing rich 2D appearance cues that enhance texture realism in the final 3D asset.

In summary, our main contributions are as follows:
\begin{itemize}
    \item We construct an effective and feed-forward framework that leverages the powerful prior of 3D generative models to enable efficient and high-quality 3D asset editing from a single edited view.
    \item We introduce Voxel FlowEdit, a voxel-flow editing mechanism. It constructs the source-to-target translation of 3D assets within the sparse voxel latent space by utilizing a specially adapted velocity field, achieving globally coherent 3D geometric deformation.
    \item We develop a dedicated normal-guided single-to-multi-view generation model which serves as an external appearance prior to overcome the limitation of compressed 3D appearance representations, restoring high-fidelity textures onto the edited geometry.
\end{itemize}
\section{Related Work}
\paragraph{3D Model Generation.}
Recent progress in 3D generative modeling has evolved from lifting 2D observations~\cite{poole2023dreamfusion,lin2023magic3d} to learning fully 3D-native representations that jointly model geometry and appearance~\cite{honglrm}.  
Early image-to-3D frameworks employed NeRF or mesh decoders to reconstruct assets from few views~\cite{mildenhall2021nerf,niemeyer2020differentiable,zhang2021nerfactor,oechsle2021unisurf}, while subsequent diffusion-based pipelines leveraged 2D priors for text-to-3D synthesis via score distillation~\cite{poole2023dreamfusion,wang2023prolificdreamer,lin2023magic3d,raj2023dreambooth3d}.  
To improve view consistency and scalability, several works proposed generating multi-view images as intermediate supervision before reconstructing 3D assets~\cite{liu2023zero,ye2024consistent,shimvdream,liu2024one}, achieving better alignment but still constrained by 2D lifting.  
More recently, large-scale 3D-native frameworks have emerged, learning structured latent spaces directly from massive 3D corpora~\cite{honglrm,tang2024lgm,yang2024hunyuan3d,zhao2025hunyuan3d,xiang2025structured,li2025controllable}.  
These models enable feed-forward generation of meshes, radiance fields, or 3D Gaussian representations conditioned on images or text, substantially advancing fidelity and controllability.  
Parallelly, autoregressive models like SAR3D~\cite{chen2024sar3d} utilize multi-scale 3D VQVAE for unified generation and understanding, while OctGPT~\cite{wei2025octgpt} leverages octree-based representations to scale up 3D shape synthesis.
\paragraph{2D Image Editing.}
Early diffusion-based image editing reconstructs a given image by inverting it into the latent space of a pretrained model and then applying localized manipulations via attention control or latent blending to realize semantic and structural changes~\cite{hertz2023prompt,kawar2023imagic,mengsdedit,mokady2023null,cao2023masactrl,yang2023paint,tumanyan2023plug,you2016image}. 
In parallel, training-based approaches adapt the generator or lightweight adapters to the target domain, improving edit fidelity and controllability through finetuning or conditioning modules such as DreamBooth~\cite{ruiz2023dreambooth,kumari2023multi}, LoRA~\cite{hu2022lora}, and ControlNet~\cite{zhang2023adding,guo2024sparsectrl}. 
In contrast to diffusion-style U-Net editors, inversion-free flow-matching formulations directly construct continuous source-target transformations in the learned velocity field, avoiding iterative inversion and aligning naturally with DiT-style architectures~\cite{lipman2023flow,kulikov2025flowedit,ni2023conditional}. Given that paired 3D editing data are largely unavailable and that the underlying generative backbone adopts a flow-matching formulation~\cite{lipman2023flow,liu2023flow}, this compatibility makes the flow-based editing paradigm well suited to our setting.
\paragraph{3D Model Editing.}
Recent 3D editing methods are characterized by how 2D guidance is coupled to the 3D representation. A first line iteratively optimizes 3D representations by supervising rendered views with 2D-edited images, typically via score-distillation losses~\cite{haque2023instruct, zhuang2023dreameditor, sella2023vox}. Subsequent work improves robustness and efficiency by synchronizing multi-view constraints or imposing geometry-aware priors during optimization~\cite{barda2024magicclay,wu2024gaussctrl,cai2024mv2mv}. A parallel direction leverages multi-view or video diffusion to produce edited view sets with stronger viewpoint coverage and spatial-temporal regularization before lifting back to 3D, enabling multi-view propagation of edits~\cite{mvedit2024}. More recently, 3D-native generative backbones have begun to explore feed-forward editing. These approaches typically rely on local mechanisms such as masked reconstruction~\cite{gao20253d} or multi-view inpainting~\cite{barda2025instant3dit} to localize changes. While these methods improve efficiency, their reliance on local masking or inpainting often limits them to textural changes or simple additions, struggling with edits that require globally coherent geometric deformation. Contemporaneous to our work, VoxHammer~\cite{li2025voxhammer} and Nano3D~\cite{ye2025nano3dtrainingfreeapproachefficient} independently explore a highly similar pipeline for feed-forward editing on 3D-native generative models. Developed in parallel, our study shares the core motivation of enabling direct manipulations within the 3D latent space. These works collectively demonstrate the timeliness and significance of shifting towards fully 3D-native feed-forward editing paradigms.

\section{Method}
\label{sec:method}
\begin{figure*}[h]  
  \centering
  \includegraphics[width=\linewidth]{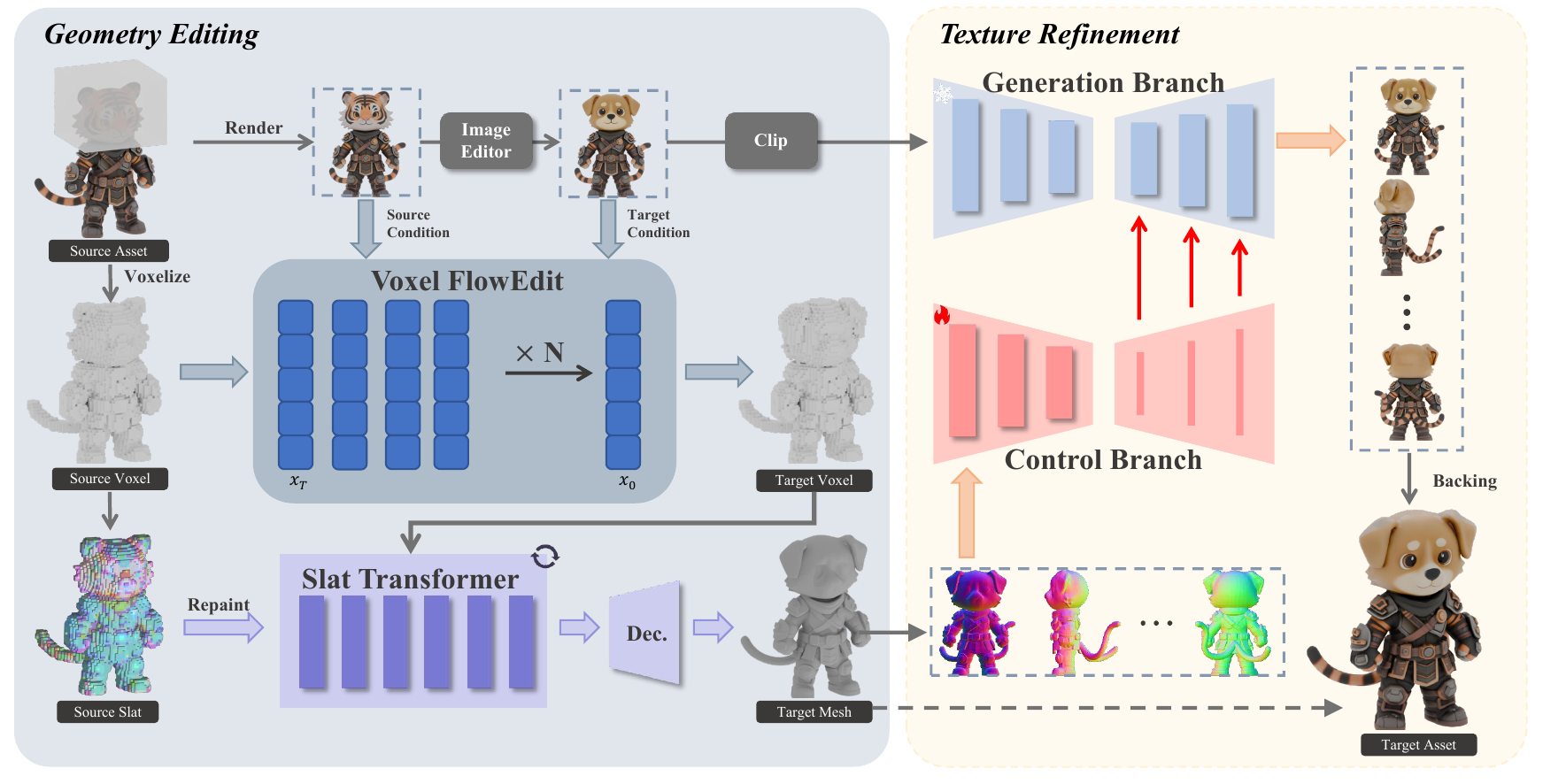}
 \caption{
Overview of Easy3E. 
The framework operates in two main stages: Geometry Editing and Texture Refinement. 
Starting from a rendered source view, an edited target image provides the guidance for editing. 
In the Geometry Editing stage, the Voxel FlowEdit algorithm transforms the source voxel structure under flow-based guidance, followed by SLAT Repainting that refines local latent features to produce the target mesh. 
The Texture Refinement stage then employs a generation branch and a normal-guided control adapter to synthesize multi-view-consistent textures, which are projected and fused onto the mesh to yield the final high-fidelity 3D asset.
}
  \label{fig:pipeline}
\end{figure*}

We introduce an effective, feed-forward framework that achieves high-fidelity 3D asset editing. Given a source 3D asset $\mathcal{A}_{\text{src}}$, a 3D region mask $\mathcal{M}$, and a target image $I^{\text{tgt}}$ obtained by editing a rendered view $I^{\text{src}}$, our method produces an edited asset with consistent geometry and appearance.

The overall pipeline is illustrated in \cref{fig:pipeline}.  We first establish our model foundation by formalizing the structured latent representation (\cref{sec:foundation}). For precise geometric editing in the 3D sparse voxel domain, we introduce a novel Flow Matching-based voxel editing algorithm (\cref{sec:geom}).  For generating new latent features that define the edited geometry and appearance while maintaining source consistency, we propose a SLAT repainting technique (\cref{sec:slat}). Finally, for enhancing visual realism, we employ a texture refinement module that improves fidelity using normal-guided multi-view images generation (\cref{sec:texture}).

\subsection{Structured Latent Representation}
\label{sec:foundation}

Our editing method operates directly in the structured 3D latent space used by TRELLIS~\cite{xiang2025structured}. 
Specifically, the Structured LATent (SLAT) representation is defined as
\[
\mathbf{Z}= \big(\mathcal{V},\, \{\mathbf{z}_{\mathbf{p}}\}_{\mathbf{p} \in \mathcal{V}}\big),
\]
where the active voxel set $\mathcal{V}$ consists of voxels that intersect the surface of the 3D mesh, and $\mathbf{z}_{\mathbf{p}}$ is a local latent feature attached to each active voxel. The local latent features are obtained by fusing multi-view image features (e.g., DINOv2) projected onto these voxels. TRELLIS uses two rectified flow transformers to respectively predict the voxel structure $\mathcal{V}$ and the local latent field $\{\mathbf{z}_{\mathbf{p}}\}$. The resulting SLAT representation can be further decoded into 3DGS, mesh, or NeRF.

TRELLIS is trained under the rectified flow framework, which learns a deterministic velocity field
\[
\frac{d\mathbf{x}}{dt}=\mathbf{v}_\theta(\mathbf{x},t),
\]
following a linear path
\[
\mathbf{x}(t) = (1-t)\mathbf{x}_0 + t\mathbf{x}_1,
\]
where \(\mathbf{x}_0\) is a clean sample from the SLAT data distribution at \(t=0\), and \(\mathbf{x}_1\) is its corresponding noise sample at \(t=1\).
This noise-to-data formulation provides the flow-based perspective on which we construct edit-driven trajectories in the structured latent space.

\subsection{Sparse Voxel Editing}
\label{sec:geom}

Our editing process begins at the structural level. The voxel structure $\mathcal{V}$ is represented as a binary occupancy grid, which is encoded by a 3D VAE into a low-dimensional continuous latent vector $\mathbf{x}$. Our goal is to propagate 2D edit signals through this latent voxel space and transform the source latent $\mathbf{x}^{\text{src}}$ into a target latent $\mathbf{x}^{\text{tgt}}$ conditioned on the target image $I^{\text{tgt}}$.

~\\
\noindent\textbf{Editing Trajectory Modeling.} Inspired by FlowEdit~\cite{kulikov2025flowedit}, we model the structural editing process as a continuous trajectory in the latent space. Let $\mathbf{x}_t$ denote the latent structural state at time $t\!\in[0,1]$, and let $\mathbf{v}_{\text{edit}}(\mathbf{x}_t, t)$ be the edit-driven velocity field. The trajectory is defined by the masked ODE
\begin{equation}
\mathrm{d}\mathbf{x}_t
= \mathcal{M}_\ell \odot \mathbf{v}_{\text{edit}}(\mathbf{x}_t, t)\,\mathrm{d}t,
\label{eq:flowedit_masked}
\end{equation}
where $\mathcal{M}_\ell$ restricts updates to the editable region. Following the rectified-flow convention, we set
\[
\mathbf{x}_{t=1} = \mathbf{x}^{\text{src}},\qquad
\mathbf{x}_{t=0} = \mathbf{x}^{\text{tgt}},
\]
so that integrating the ODE traces a continuous path from the source latent structure to the desired edited one.

We construct $\mathbf{v}_{\text{edit}}$ by differencing the flow trajectories of the pretrained model under the source and target conditions. For each condition $c \in \{\text{src},\,\text{tgt}\}$, the rectified-flow formulation specifies a linear latent path
\[
\mathbf{x}^c_t = (1-t)\,\mathbf{x}^c_0 + t\,\mathbf{x}^c_1,
\]
connecting a clean endpoint $\mathbf{x}^c_0$ and a noisy endpoint $\mathbf{x}^c_1$. The velocity network satisfies
\[
\frac{d\mathbf{x}^c_t}{dt} = \mathbf{v}_\theta(\mathbf{x}^c_t, t \mid I^{c}).
\]

We impose a shared terminal noise state $\mathbf{x}^{\text{src}}_1=\mathbf{x}^{\text{tgt}}_1$ and construct an interpolated path that matches our ODE boundary conditions:
\[
\mathbf{x}_t
= \mathbf{x}^{\text{src}}_0 - \mathbf{x}^{\text{src}}_t + \mathbf{x}^{\text{tgt}}_t.
\]

Differentiating this path yields the edit velocity:
\begin{equation}
\mathbf{v}_{\text{edit}}(\mathbf{x}_t,t)
= \mathbf{v}_\theta(\mathbf{x}^{\text{tgt}}_t, t \mid I^{\text{tgt}})
 - \mathbf{v}_\theta(\mathbf{x}^{\text{src}}_t, t \mid I^{\text{src}}).
\end{equation}

\begin{figure}[t]  
  \centering
  \includegraphics[width=\linewidth]{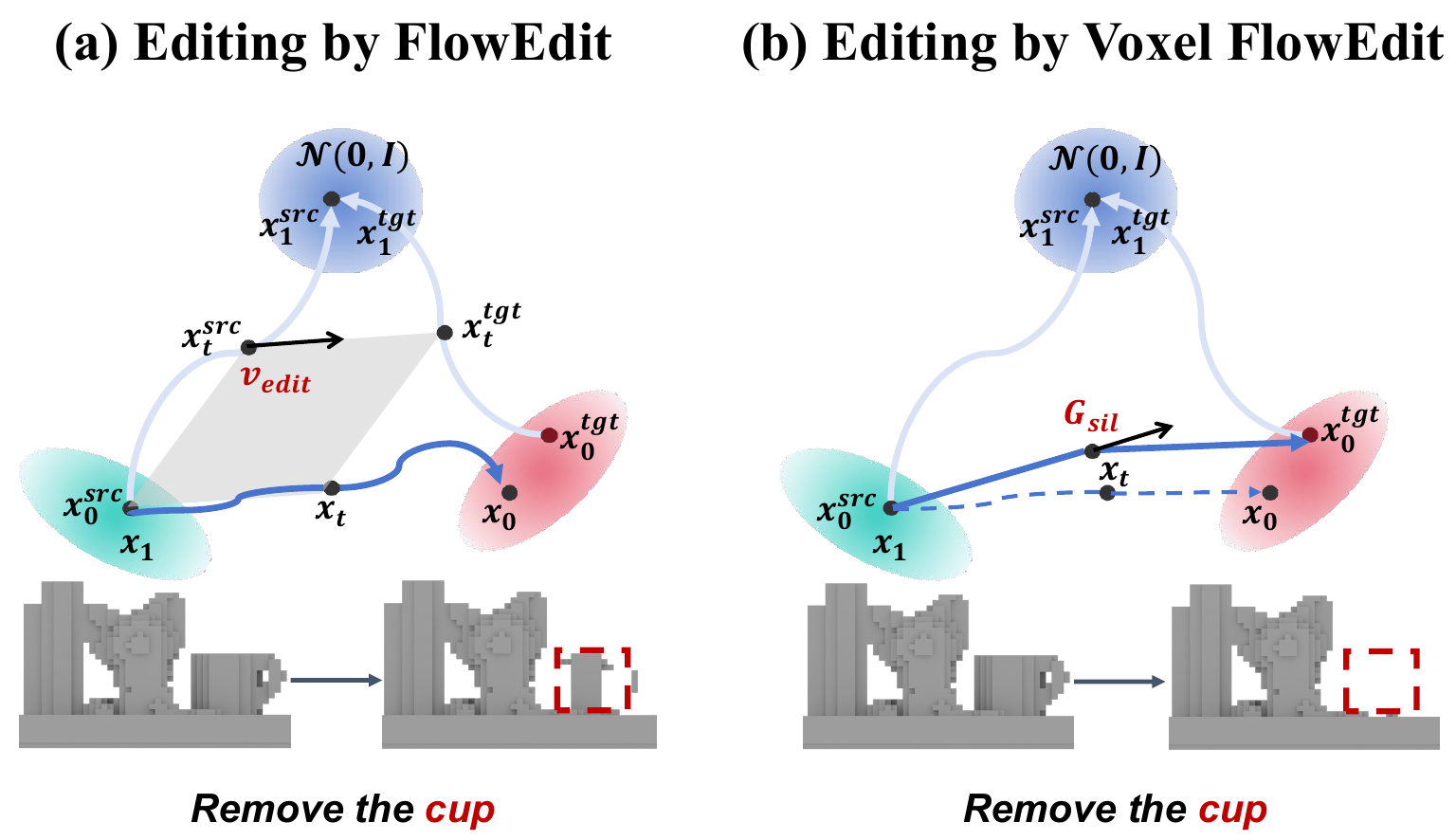}
\caption{Comparison of FlowEdit's limitations and the Voxel FlowEdit solution.
(a) Base FlowEdit: The semantic velocity $\mathbf{v}_{\text{edit}}$ is corrupted by accumulated approximation error, causing the trajectory to drift and resulting in structural corruption (red dashed box).
(b) Voxel FlowEdit: The edit is driven by external gradient guidance $\mathbf{G}_{\text{sil}}$, while internal correction $\boldsymbol{\xi}_{\text{traj}}$ maintains manifold consistency. This combined approach achieves a clean and structurally integral edit.}
  \label{fig:method1}
\end{figure}

\noindent The geometric definition of this vector is visualized in Figure \ref{fig:method1}(a).

~\\
\noindent\textbf{Guided Flow Regularization.} The base ODE in Eq.~\ref{eq:flowedit_masked} provides an efficient initialization for 3D structural editing. 
Following FlowEdit~\cite{kulikov2025flowedit}, the editing velocity $\mathbf{v}_{\text{edit}}$ is estimated by averaging conditional velocity differences over multiple noise samples.

However, this baseline often over-preserves the source structure and fails to fully reach the target edit (Fig.~\ref{fig:method1}(a)), likely due to discretization errors that push the trajectory off the data manifold. 
To mitigate this drift, we introduce auxiliary guidance terms that refine the evolution while preserving the underlying flow dynamics.

We introduce a silhouette-guidance term based on the energy
\begin{equation}
\begin{aligned}
\mathcal{E}_{\text{sil}}(\mathbf{x})
&= \operatorname{BCE}\!\big(S(\mathbf{x}),\,M_{\text{sil}}\big),\\
S(\mathbf{x})
&= 1 - \exp\!\Big(-\kappa \sum_{z} p_z(\mathbf{x})\Big),
\end{aligned}
\label{eq:silhouette_energy}
\end{equation}
where $\mathrm{BCE}$ denotes binary cross-entropy,  
$p_z(\mathbf{x})$ is the decoded voxel occupancy probability at depth $z$, and  
$S(\mathbf{x})$ is the rendered 2D silhouette obtained by accumulating occupancy along each camera ray.  
The target silhouette $M_{\text{sil}}$ is extracted from the edited target view $I^{\text{tgt}}$, and $\kappa$ controls its sharpness.
The guidance term
\[
\mathbf{G}_{\text{sil}}(\mathbf{x}_t)
= -\nabla_{\mathbf{x}}\mathcal{E}_{\text{sil}}(\mathbf{x}_t)
\]
encourages the evolving structure to match the target silhouette.

Directly applying $\mathbf{G}_{\text{sil}}$ may push the edited state off the smooth flow manifold. 
To counteract this, we introduce a trajectory-consistency correction 
$\boldsymbol{\xi}_{\text{traj}}(\mathbf{x}_t)$~\cite{kulikov2025flowedit}, 
which projects the perturbed latent state back onto the interpolating manifold:
\[
\boldsymbol{\xi}_{\text{traj}}(\mathbf{x}_t)
= \widehat{\mathbf{x}}^{\text{tgt}}_{0|t}
 - \widehat{\mathbf{x}}^{\text{src}}_{0|t},
\]
where the clean-state estimates are obtained by back-projecting each trajectory,
\[
\widehat{\mathbf{x}}^{c}_{0|t}
= \mathbf{x}^c_t - t\,\mathbf{v}_\theta(\mathbf{x}^c_t, t \mid c),
\qquad c\in\{\text{src},\text{tgt}\}.
\]

Combining the semantic velocity $\mathbf{v}_{\text{edit}}$ with these auxiliary terms yields the controllable flow-regularized update:
\begin{equation}
\begin{aligned}
\mathrm{d}\mathbf{x}_t
&= \mathcal{M}_\ell \odot \mathbf{v}_{\text{edit}}(\mathbf{x}_t,t)\,\mathrm{d}t \\
&\quad +\, \mathcal{M}_\ell \odot 
\Big(
\Gamma\,\boldsymbol{\xi}_{\text{traj}}(\mathbf{x}_t)
- \eta\,\mathbf{G}_{\text{sil}}(\mathbf{x}_t)
\Big)\,\mathrm{d}t,
\end{aligned}
\label{eq:final_masked_ode}
\end{equation}
where the constants $\Gamma$ and $\eta$ control the relative strength of the trajectory correction and gradient guidance.
\noindent The successful edit and stable trajectory achieved by this flow-regularized update are visualized in \cref{fig:method1}(b).

To obtain a stable discrete realization, we integrate the above flow in multiple steps from $t{=}1$ to $0$. 
Our final algorithm is summarized in \cref{alg:flowedit}.
\begin{algorithm}[t]
\caption{Sparse Voxel FlowEdit}
\label{alg:flowedit}
\begin{algorithmic}
\State \textbf{Input:} source latent $\mathbf{x}^{\text{src}}_0$, $\{t_i\}_{i=0}^T$, $\mathcal{M}_\ell$, $\Gamma,\eta$
\State \textbf{Output:} edited latent $\mathbf{x}^{\text{tgt}}_0$
\State \textbf{Init:} $\mathbf{x}_{t_T} \gets \mathbf{x}^{\text{src}}_0$
\For{$i = T$ down to $1$}
  \State Sample $\boldsymbol{\epsilon}_{t_i} \sim \mathcal{N}(\mathbf{0},\mathbf{I})$
  \State $\mathbf{x}^{\text{src}}_{t_i} \gets (1 - t_i)\,\mathbf{x}^{\text{src}}_0 + t_i\,\boldsymbol{\epsilon}_{t_i}$
  \State $\mathbf{x}^{\text{tgt}}_{t_i} \gets \mathbf{x}^{\text{src}}_{t_i} + \mathbf{x}_{t_i} - \mathbf{x}^{\text{src}}_0$ 
  \State $\tilde{\mathbf{x}}_{t_{i-1}} \gets 
    \mathbf{x}_{t_i} + \Delta t\,\mathcal{M}_\ell \odot \mathbf{v}_{\text{edit}}(\mathbf{x}_{t_i}, t_i)$
  \State $\widehat{\mathbf{x}}^{i}_{0|t_i} \gets \mathbf{x}^{i}_{t_i} - t_i\,\mathbf{v}_\theta(\mathbf{x}^{i}_{t_i}, t_i \mid I^{i}),\quad i \in \{\text{src},\,\text{tgt}\}$
  \State $\boldsymbol{\xi}_{\text{traj}}(\mathbf{x}_{t_i}) \gets 
    \widehat{\mathbf{x}}^{\text{tgt}}_{0|t_i} - \widehat{\mathbf{x}}^{\text{src}}_{0|t_i}$
    \State $\mathbf{G}^{\text{sil}}_{t_{i-1}} \gets 
    \mathbf{G}_{\text{sil}}(\tilde{\mathbf{x}}_{t_{i-1}})$
  \State $\mathbf{x}_{t_{i-1}} \gets \tilde{\mathbf{x}}_{t_{i-1}} 
    + \Delta t\,\mathcal{M}_\ell \odot
    \big(\Gamma\,\boldsymbol{\xi}_{\text{traj}}(\mathbf{x}_{t_i}) - \eta\,\mathbf{G}^{\text{sil}}_{t_{i-1}}\big)$
    \EndFor
\State \textbf{Return:} $\mathbf{x}^{\text{tgt}}_0 \gets \mathbf{x}_{t_0}$
\end{algorithmic}
\end{algorithm}

\subsection{SLAT Repainting}
\label{sec:slat}

To further refine local geometry and appearance beyond the sparse structural edits, 
we introduce a latent-level repainting stage that updates voxel features in the editable region while preserving the source characteristics elsewhere.

Building upon the edited sparse voxels $\mathcal{V}_{\text{tgt}}$ obtained from Voxel FlowEdit (\cref{sec:geom}), 
we refine local geometry by updating the latent feature vectors $\{\mathbf{z}_{\mathbf{p}}\}_{\mathbf{p} \in \mathcal{V}_{\text{tgt}}}$ within the editable region, 
while anchoring the unedited ones to the source distribution. Let $\mathcal{M}_z$ denote the per-latent edit mask derived from the mesh-space mask $\mathcal{M}$.

Note that $\mathbf{v}_{\theta}$ here operates on the local latent features $\mathbf{z}$. 
At each discrete step $k$, the latent update follows a repainting process:
\begin{equation}
\begin{aligned}
\mathbf{z}_{k-1}
&= \mathcal{M}_z\odot\Big[\mathbf{z}_{k}
+\Delta t\,\mathbf{v}_\theta(\mathbf{z}_{k},t_k \mid I^{\text{tgt}})\Big] \\
&\quad +(1-\mathcal{M}_z)\odot\Big[(1-t_{k-1})\,\mathbf{z}^{\text{src}}+t_{k-1}\,\boldsymbol{\epsilon}_{k-1}\Big],
\end{aligned}
\label{eq:local-latent-repaint}
\end{equation}
where $\mathbf{z}^{\text{src}}$ is the initial source latent vector at $t_0$, 
and $\boldsymbol{\epsilon}_k\!\sim\!\mathcal{N}(\mathbf{0},\mathbf{I})$ denotes Gaussian noise.

The two masked terms jointly ensure local refinement and global preservation:  
the first term applies target-conditioned velocity for structural refinements in the editable region, 
whereas the second term replays the forward-diffused source trajectory to maintain global appearance and geometry. 
In practice, a softly feathered mask $\widetilde{\mathcal{M}_z}=\operatorname{blur}(\mathcal{M}_z;\sigma_b)$ is used to prevent seam artifacts.

After reaching $k=0$, the final edited latent field 
$\mathbf{Z}= \big(\mathcal{V}_{\text{tgt}},\, \{\mathbf{z}_{\mathbf{p}}\}_{\mathbf{p} \in \mathcal{V}_{\text{tgt}}}\big)$ 
is decoded by TRELLIS to produce the refined 3D mesh.

\subsection{Texture Refinement}
\label{sec:texture}
Given the edited 3D mesh decoded from the preceding stages, we optionally apply a texture refinement module to enhance appearance realism. 
~\\
~\\
\noindent \textbf{Control Branch.} The Control Branch serves to inject precise geometric guidance for the subsequent multi-view synthesis. It is constructed by combining a frozen ControlNet~\cite{zhang2023adding} with a trainable Ctrl-Adapter~\cite{lin2025ctrl}. The ControlNet receives per-view normal maps $\{\mathbf{N}_v\}$ rendered from the edited geometry and extracts multi-scale spatial features that encode local surface cues. The Ctrl-Adapter then learns to align and inject these control features into the main generative network, effectively acting as a geometry-aware conditioning module. This design enables precise and efficient control over texture synthesis guided strictly by the edited 3D shape.
~\\
~\\
\noindent \textbf{Generation Branch.} Conditioned on the geometric guidance extracted by the Control Branch, we adopt the multiview-diffusion architecture from ERA3D~\cite{li2024era3d} to synthesize multi-view images consistent with the guiding geometry. The generation backbone takes the edited image $I^{\text{tgt}}$ as primary context, using the control features $\mathbf{C}$ to ensure the output is consistent with the edited geometry. This network synthesizes six geometry-consistent auxiliary views $\{I'_v\}_{v=1}^{6}$ under predefined camera poses. This synthesis process propagates fine appearance details onto the novel viewpoints, providing reliable texture information for the following fusion stage. During training, only the parameters of the Ctrl-Adapter are updated to learn the feature alignment, while the ControlNet and the Era3D backbone remain frozen.
~\\
~\\
\noindent \textbf{Texture Fusion.} The final generated appearance is transferred back to the UV space via a robust fusion process. The synthesized views $\{I'_v\}$ are projected onto the 3D mesh and fused into the final UV texture $\mathbf{T}$ using a visibility-aware, mask-weighted blending scheme. This process uses the softly feathered edit mask $\widetilde{\mathcal{M}}$ to prioritize integration in the edited regions, while strictly retaining the original appearance in unedited areas. This efficient transfer yields significantly sharper and more realistic textures for the edited assets.

\section{Experiments}
\label{sec:experiment}
\subsection{Implementation Details}
For Voxel FlowEdit, we adopt a target-side classifier-free guidance (CFG) scale between 5 and 15,  
while the source-side CFG is fixed to 5.  
The latent ODE is discretized by 25 sampling steps,  
and the edit velocity $\mathbf{v}_{\text{edit}}$ is computed by averaging over $n_{\text{avg}}\!\in\{2,4\}$  
to improve stability.  
For the regularization terms, we normalize the silhouette-guidance gradient so that its $\ell_2$ norm matches 
that of $\mathbf{v}_{\text{edit}}$, allowing a weighting coefficient of 0.2 for the silhouette term.  The trajectory-consistency residual is weighted by 0.1.

The normal-guided Ctrl-Adapter is trained on a subset of Objaverse~\cite{deitke2023objaverse},  
where each asset is rendered into six views of size $512{\times}512$ with corresponding normal maps.  
The adapter is trained to synthesize auxiliary views that guide the texture refinement stage.

We construct our evaluation set using 100 3D assets collected from Sketchfab Website, the NPHM~\cite{giebenhain2023learning} dataset for real human heads, the THuman2.1~\cite{tao2021function4d} dataset for clothed human bodies, and the Objaverse~\cite{deitke2023objaverse} dataset for objects.

\subsection{Comparisons}
\paragraph{Baselines.}
\begin{figure*}[h]
  \centering
  \includegraphics[width=\linewidth]{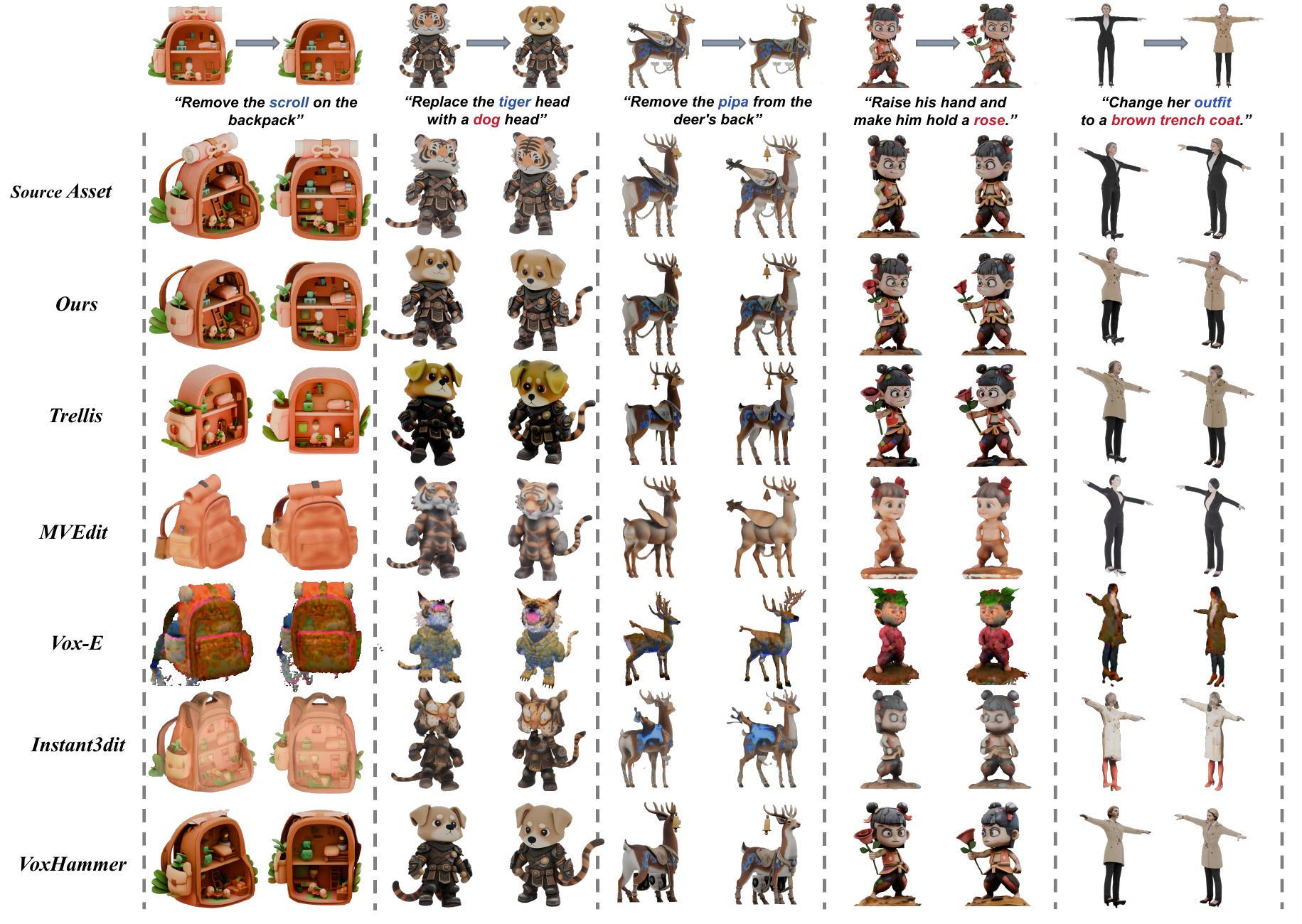}
  \caption{
    Qualitative comparison. Our method achieves clean geometry and consistent appearance across multiple views, faithfully realizing the target edits while strictly preserving unedited regions. Competing methods either retain the original geometry (MVEdit) or exhibit strong structural distortion and inconsistency (Vox-E, Instant3dit). Although VoxHammer also utilizes mask inputs to guide the editing process, it struggles to maintain high fidelity outside the masked areas.
  }
  \label{fig:qualitative_comparison}
\end{figure*}

We compare our method against several representative works.
(1) TRELLIS~\cite{xiang2025structured}, which serves as our generative backbone. We feed the edited target image $I_{\text{tgt}}$ directly into its 3D generation pipeline and evaluate its ability to reproduce the edit while preserving consistency with the original 3D model.  
(2) MVEdit~\cite{mvedit2024}, a multi-view diffusion framework that uses a training-free 3D adapter to jointly denoise rendered views and output textured meshes.  
(3) Vox-E~\cite{sella2023vox}, a voxel-based volumetric editing method that learns a volumetric representation from oriented images and edits existing 3D objects under diffusion priors.  
(4) Instant3dit~\cite{barda2025instant3dit}, a feed-forward multiview inpainting framework for fast 3D editing via view generation and reconstruction. (5) VoxHammer~\cite{li2025voxhammer}, a feed-forward 3D editing framework built upon TRELLIS that uses an image and a mask to guide the editing process.
Since methods (2)-(4) are primarily text-guided frameworks rather than direct image-driven 3D generation, we provide them with a unified text prompt that semantically describes our image-based edit to ensure a fair comparison.
~\\
~\\
\noindent \textbf{Qualitative Comparison.} As shown in \cref{fig:qualitative_comparison}, our method produces clean, view-consistent edits while preserving the source asset's global structure. Compared to TRELLIS and VoxHammer, our framework avoids overfitting and better maintains identity consistency with the original 3D model. MVEdit generates plausible textures but introduces negligible geometric changes and view-wise inconsistency. Both Vox-E and Instant3dit fail to maintain structural integrity under complex edits, with Vox-E also requiring significantly longer inference time. Overall, our approach achieves a superior balance of edit controllability, identity preservation, and visual fidelity.

~\\
\noindent \textbf{Quantitative Comparison.} We evaluate performance using four widely adopted quantitative metrics: CLIP-T (text-image alignment)~\cite{radford2021learning}, DINO-I (perceptual quality)~\cite{caron2021emerging}, LPIPS (perceptual similarity)~\cite{zhang2018unreasonable}, and FID (distributional realism)~\cite{heusel2017gans}. As shown in \cref{tab:quantitative}, our method consistently achieves the best performance across all four metrics. These results confirm that our framework produces higher overall 3D quality and text alignment than representative baselines, yielding more coherent, high-fidelity 3D assets.

\begin{table}[t]
\centering
\caption{Quantitative comparison on the 3D editing benchmark. 
Higher is better for CLIP-T, and DINO-I; 
lower is better for LPIPS and FID. 
Our method consistently achieves the best performance across all metrics.}
\label{tab:quantitative}
\resizebox{\linewidth}{!}{
\begin{tabular}{lcccc}
\toprule
Method & CLIP-T~$\uparrow$ & DINO-I~$\uparrow$ & LPIPS~$\downarrow$ & FID~$\downarrow$ \\
\midrule
TRELLIS~\cite{xiang2025structured} & 0.323 & 0.895 & 0.243 & 30.2 \\
MVEdit~\cite{mvedit2024} & 0.267 & 0.851 & 0.282 & 67.6 \\
Vox-E~\cite{sella2023vox} & 0.266 & 0.734 & 0.673 & 90.3 \\
Instant3DiT~\cite{barda2025instant3dit} & 0.285 & 0.874 & 0.286 & 49.7 \\
VoxHammer~\cite{li2025voxhammer} & 0.295 & 0.938 & 0.193 & 31.7\\
\midrule
\textbf{Ours} & \textbf{0.326} & \textbf{0.952} & \textbf{0.138} & \textbf{25.8} \\
\bottomrule
\end{tabular}
}
\end{table}

~\\
\noindent\textbf{User Study.}
 We also conduct a user study to further quantitatively validate our method. Participants were asked to view the editing prompt, alongside videos rendered from both the source 3D assets and the 3D assets edited by our method and four competing methods, and then respond to a series of questions: 

\begin{table}[t]
\centering
\caption{User study results. Percentage of times each question is rated best (higher is better).}
\label{tab:userstudy}
\resizebox{\linewidth}{!}{
\begin{tabular}{lccccc}
\toprule
Question & Q1 $\uparrow$ & Q2 $\uparrow$ & Q3 $\uparrow$ & Q4 $\uparrow$ & Q5 $\uparrow$ \\
\midrule
\textbf{Ours}                        & 88.98 & 94.63 & 94.92 & 97.51 & 97.00 \\
\bottomrule
\end{tabular}
}
\end{table}

\begin{itemize}
    \item \textbf{Q1:} Which method best follows the given input prompt? (\emph{Prompt Preservation})
    \item \textbf{Q2:} Which method best retains the geometry and texture of the unedited regions? (\emph{Identity Preservation})
    \item \textbf{Q3:} Which method best produces edited geometry and texture? (\emph{3D Editing Quality})
    \item \textbf{Q4:} Which method best maintains 3D consistency? (\emph{3D Consistency})
    \item \textbf{Q5:} Which method is best overall considering the above four aspects? (\emph{Overall})
\end{itemize}

We collected statistics from 46 participants across 10 groups of editing results. For a fair comparison, the video results for each case were randomly shuffled. As shown in \cref{tab:userstudy}, our method remarkably outperforms other methods in prompt preservation, identity preservation, 3D editing quality, and 3D consistency, and is rated as the best in overall quality. These results demonstrate that our method is highly favored by users, highlighting its effectiveness across various editing dimensions.

\subsection{Ablation Studies}
\label{sec:ablation}

\begin{figure}[t]
  \centering
  \includegraphics[width=\linewidth]{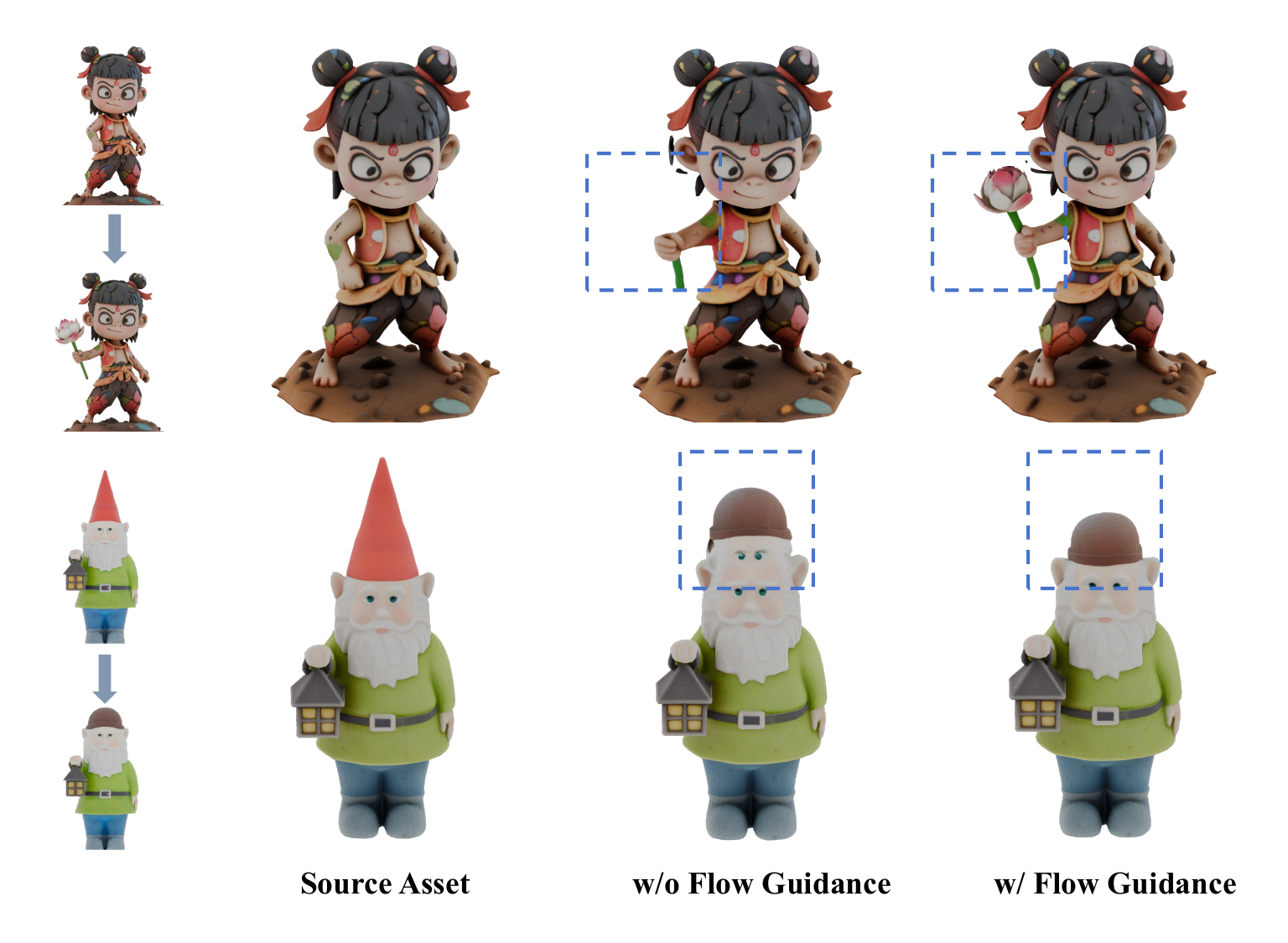}
  \caption{
  Comparison between without and with Flow Guidance. Disabling $\mathbf{G}_{\text{sil}}$ and $\boldsymbol{\xi}_{\text{traj}}$ jointly leads to structural collapse and view-inconsistent deformation, whereas enabling both yields stable and silhouette-aligned geometry.
  }
  \label{fig:ablation_flow}
\end{figure}

\noindent \textbf{Guided Flow Regularization.}
We ablate the auxiliary guidance terms in \cref{eq:final_masked_ode} to assess their impact on structural stability. Specifically, we compare the base ODE update without auxiliary guidance against the full guided formulation with both the silhouette gradient and the trajectory correction enabled. We toggle these two terms jointly, since $\mathbf{G}_{\text{sil}}$ aligns the evolving structure with the target silhouette, while $\boldsymbol{\xi}_{\text{traj}}$ regularizes the dynamics by projecting $\mathbf{x}_t$ back onto the flow manifold. Using only one of them leads to unbalanced updates and unstable geometry. As shown in \cref{fig:ablation_flow}, removing both terms results in a combination of over-preservation and structural drift, whereas enabling both yields coherent and well-aligned deformations.

\begin{figure}[t]
  \centering
  \includegraphics[width=\linewidth]{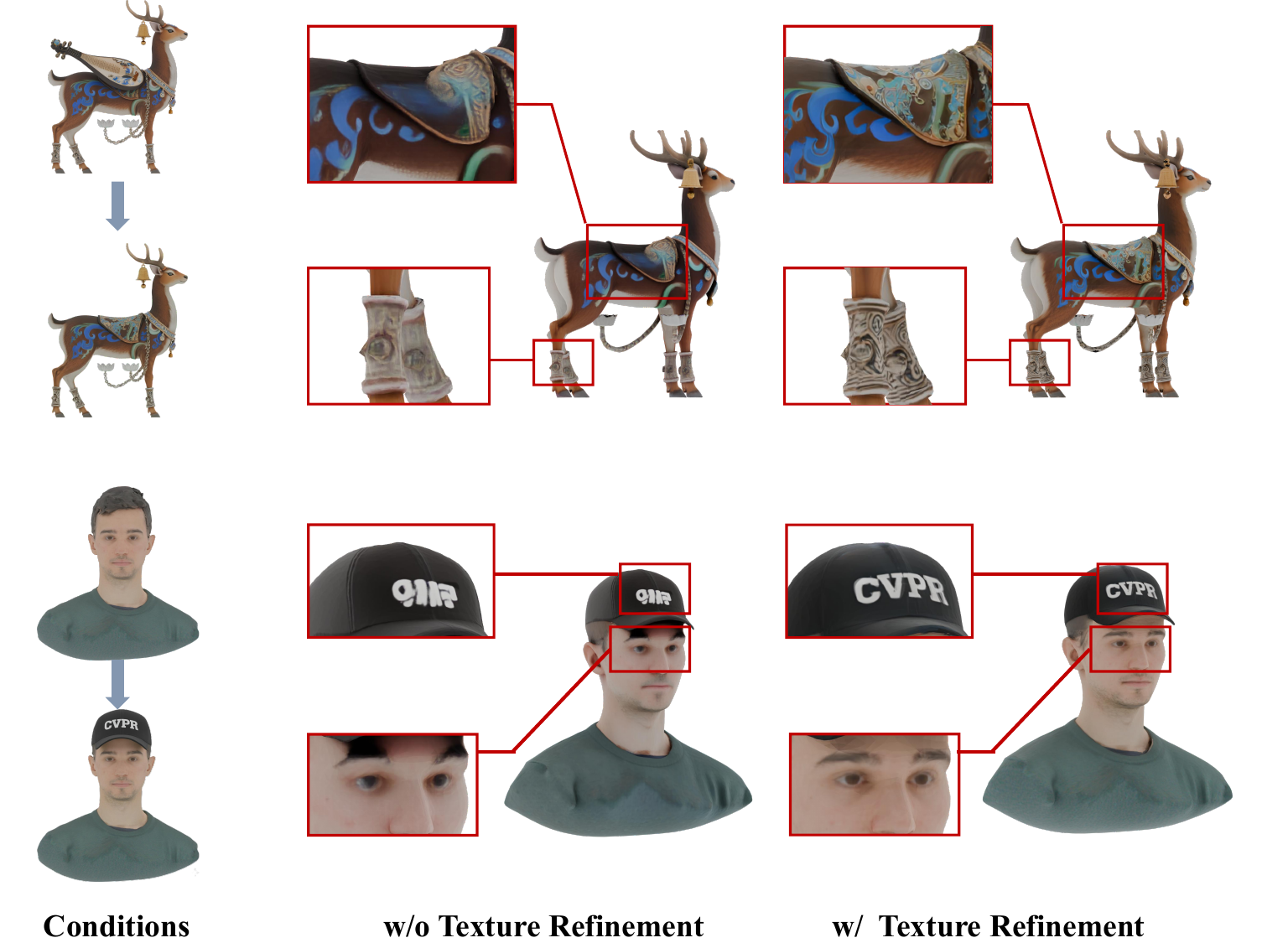}
  \caption{
  Comparison between without and with Texture Refinement. The refinement stage significantly enhances surface detail and view-consistent appearance.
  }
  \label{fig:ablation_texture}
\end{figure}

\noindent \textbf{Texture Refinement.} We further investigate the effect of the normal-guided appearance refinement module. This component restores high-frequency texture details and harmonizes lighting across views via normal-conditioned feature modulation. Without Texture Refinement, the edited regions become noticeably blurrier and exhibit color bias, as illustrated in \cref{fig:ablation_texture}.

\section{Conclusion}

We have presented a unified feed-forward framework for 3D asset editing that integrates geometric transformation, latent-space refinement, and texture enhancement within a single generative pipeline. Our method builds upon the 3D-native TRELLIS representation, enabling coherent large-scale deformation and fine-grained texture editing directly from a single-view input. By combining Voxel FlowEdit, SLAT repainting, and normal-guided texture refinement, the proposed system effectively bridges 2D editing flexibility with 3D consistency and realism. Comprehensive experiments demonstrate that our approach achieves stable geometry, faithful texture preservation, and visually consistent results across diverse assets, offering a practical paradigm for efficient and controllable 3D editing.

~\\
\noindent \textbf{Limitations.} While our framework delivers robust and realistic edits, its performance remains bounded by the generative capacity of TRELLIS, particularly under extreme geometric modifications. Moreover, the normal-guided refinement currently operates on relatively low-resolution synthesized views, which would limit the recovery of very fine textures. We believe these limitations can be mitigated in future work through higher-resolution generation and stronger geometric priors.

{
    \small
    \bibliographystyle{ieeenat_fullname}
    \bibliography{main}
}
\appendix
\clearpage
\setcounter{page}{1}
\maketitlesupplementary

\section{Editing Efficiency}

\begin{table}[htbp]
\centering
\caption{Runtime comparison with different methods.}
\label{tab:edit_speed}
\begin{tabularx}{\linewidth}{>{\centering\arraybackslash}X|>{\centering\arraybackslash}X}
\toprule
Method & Runtime \\
\midrule
Vox-E~\cite{sella2023vox} & 37 min \\
MVEdit~\cite{mvedit2024} & 212 s \\
Instant3dit~\cite{barda2025instant3dit} & 25 s \\
\textbf{Ours} & \textbf{75 s} \\
\bottomrule
\end{tabularx}
\end{table}

We compare the efficiency of our approach with three baselines: Instant3dit, MVEdit, and Vox-E, as shown in \cref{tab:edit_speed}. Instant3dit is the fastest (25 seconds) due to its highly streamlined pipeline, but this compactness often limits its ability to handle complex disentanglement, resulting in lower fidelity. In contrast, MVEdit incurs a significantly higher computational cost of 212 seconds, as it relies on heavy iterative multi-view diffusion refinement. Vox-E is even more time-consuming, requiring full 3D optimization of the voxel grid, which takes approximately 37 minutes per edit. Our method operates in a sweet spot with a total runtime of 75 seconds. Adopting an efficient feed-forward design similar to Instant3dit, our approach eliminates the need for time-consuming per-scene optimization. However, unlike the unified pipeline of Instant3dit, we utilize a structured workflow decomposed into geometry editing (30 seconds), texture refinement (30 seconds), and back-projection (15 seconds). This 75 seconds duration allows us to achieve substantially better consistency and detail than Instant3dit, while remaining orders of magnitude faster than the optimization-heavy baselines.

\section{Effectiveness of Texture Refinement}

\begin{figure}[t]
    \centering
    \includegraphics[width=\linewidth]{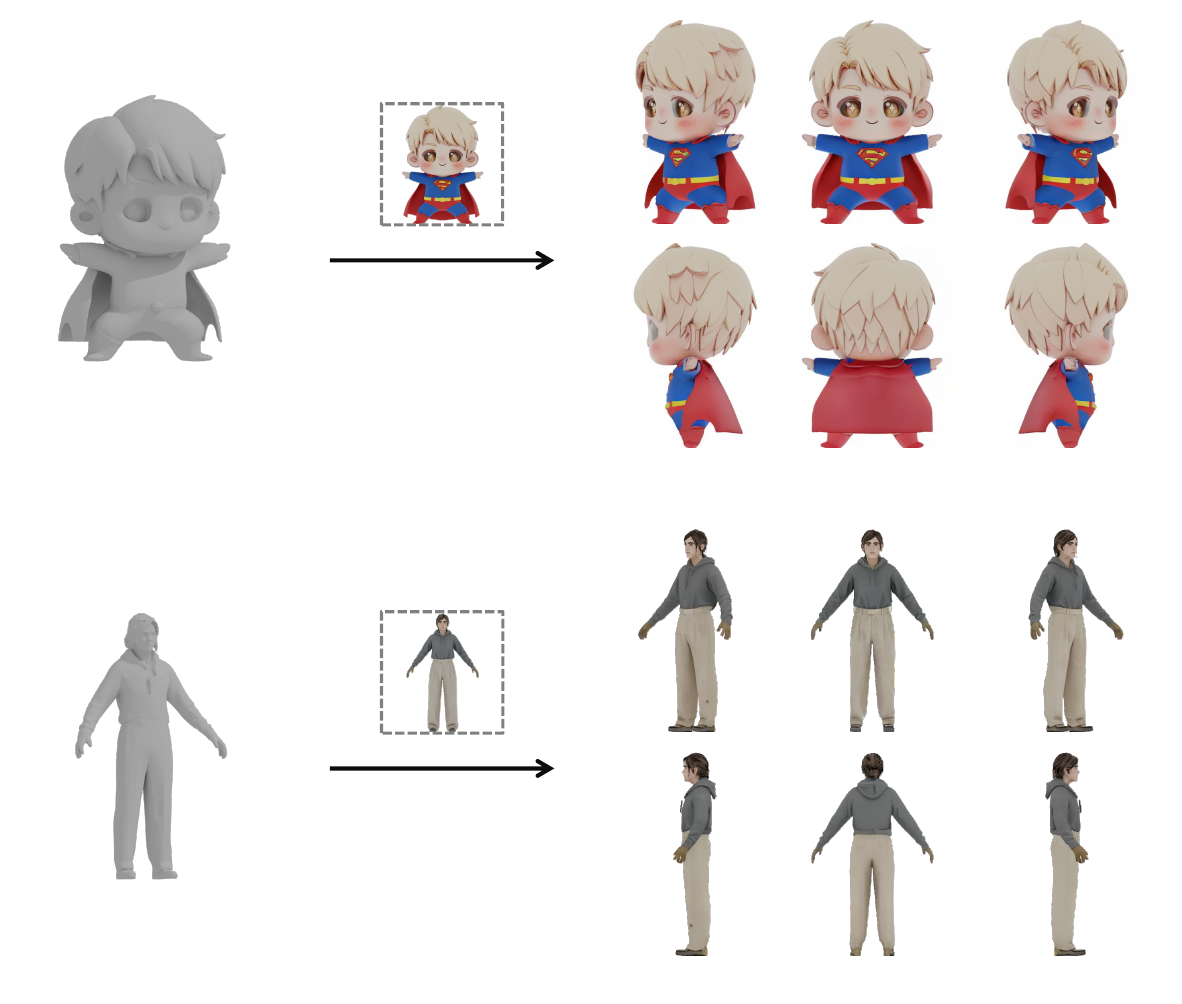} 
    \caption{Visualizations of the normal-guided multi-view diffusion module. Taking rendered normal maps and a reference image as input, the module generates multi-view images that are both texture-rich and geometrically consistent, serving as robust priors for 3D texture recovery.}
    \label{fig:mv_results}
\end{figure}

\cref{fig:mv_results} presents the qualitative results obtained from our normal-guided Multi-view Diffusion Module. Conditioned on multi-view normal maps rendered from the input mesh and a single reference image, the module synthesizes a coherent sequence of multi-view images. As shown in the visualization, the generated images exhibit high-fidelity textures with intricate details. More importantly, benefiting from the structural guidance of surface normals, the results demonstrate rigorous cross-view consistency, where the object identity and geometric features remain stable across varying camera poses. This ensures that the subsequent texture back-projection step produces a seamless 3D model without alignment artifacts.

\section{More Results}

\begin{figure*}[t]
  \centering
  \includegraphics[width=\linewidth]{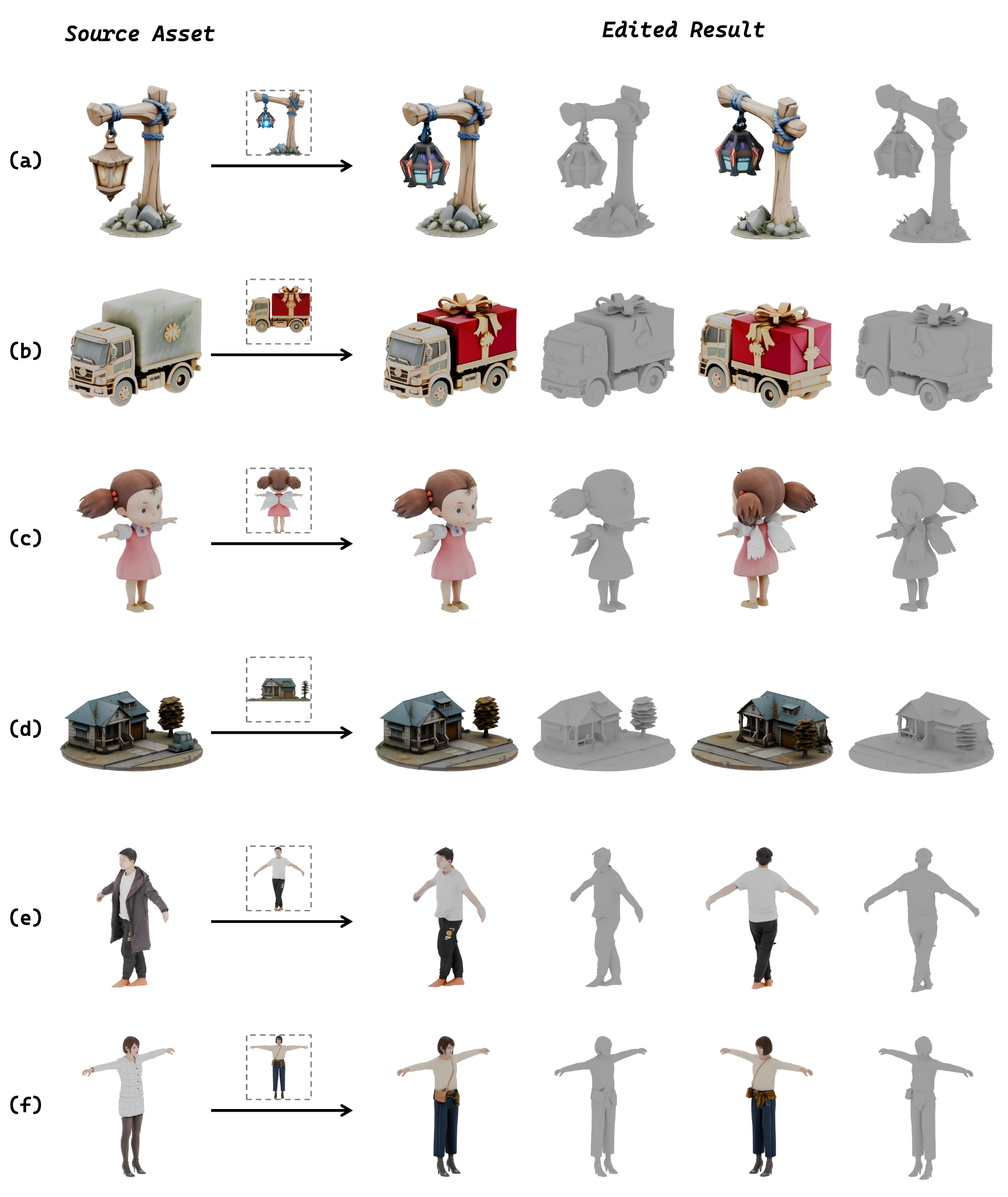}
  \caption{More visualization results.}
  \label{fig:more_results}
\end{figure*}

We present six supplementary examples in \cref{fig:more_results} to further validate the robustness of our approach. (a)-(c) demonstrate our capabilities on non-realistic objects; observe that the edited regions undergo significant geometric deformation, while the geometry of the unedited regions is strictly preserved. (d) illustrates the effectiveness of our method in scene-level editing. (e)-(f) showcase results on human subjects, where the clothing is successfully modified with high visual quality while maintaining the texture fidelity of the unedited body parts.

\end{document}